\documentclass{llncs}
\usepackage{makeidx}  % allows for indexgeneration
\usepackage[latin1]{inputenc} 
\usepackage{graphicx}
\usepackage{url}

   \def\YaTeA{{\rmfamily Y\kern-.36em%
    \lower.7ex\hbox{A}\kern-.25em%
    T\kern-.1667em\lower.7ex\hbox{E}\kern-.08emA}~}

\begin{document}
\frontmatter          % for the preliminaries
\pagestyle{headings}  % switches on printing of running heads
%\addtocmark{Hamiltonian Mechanics} % additional mark in the TOC
%
%\mainmatter              % start of the contributions
%
%%%%%%%%
%
% Which contribution of the terminological resources in the term
% extraction process?
%
%%%%%%%%
%
% Contribution of the terminological resources in the term
% extraction process
%
%%%%%%%%
%
% Recycling terminological resources in a term extraction process
%
%%%%%%%%
%
% Improving term extraction with terminological resources
%
%%%%%%%%
%
% Improvement of the term extraction process with terminological resources
%
%%%%%%%%
%
% Using exogenous information in an endogenous term extraction:
% contribution of terminological resources
%

\title{Improving term extraction with terminological resources}
\titlerunning{}  % abbreviated title (for running head)
\author{Sophie Aubin \and Thierry Hamon}

\authorrunning{Aubin and Hamon}   % abbreviated author list (for running head)
%
%%%% modified list of authors for the TOC (add the affiliations)
\tocauthor{Sophie Aubin (Université Paris Nord),
Thierry Hamon (Université Paris Nord)}
\institute{LIPN -- UMR CNRS 7030\\
  99 av. J.B. Clément, F-93430 Villetaneuse\\
  Tél. : 33 1 49 40 40 82, Fax. : 33 1 48 26 07 12\\
  \email{firstname.lastname@lipn.univ-paris13.fr},\\
  WWW home page: www-lipn.univ-paris13.fr/\texttt{\~}lastname}
% Princeton University, Princeton NJ 08544, USA,\\
% \email{I.Ekeland@princeton.edu},\\ WWW home page:
% \texttt{http://users/\homedir iekeland/web/welcome.html}
% \and
% Universit\'{e} de Paris-Sud,
% Laboratoire d'Analyse Num\'{e}rique, B\^{a}timent 425,\\
% F-91405 Orsay Cedex, France}

\maketitle              % typeset the title of the contribution

\begin{abstract}
% The abstract should summarize the contents of the paper
% using at least 70 and at most 150 words. It will be set in 9-point
% font size and be inset 1.0 cm from the right and left margins.
% There will be two blank lines before and after the Abstract. \dots
Studies of different term extractors on a corpus of the biomedical
domain revealed decreasing performances when applied to highly
technical texts. % they lack portability to highly technical domains 
% because of the difficulty
% or impossibility of customising linguistic data. 
% The difficulty or impossibility of customising them to new domains is
% an additional limitation. 
% In this paper, we propose to use external terminologies to influence
% generic linguistic data in order to augment the
% quality of the extraction. 
Facing the difficulty or impossibility to customize existing tools, we
developed a tunable term extractor. It exploits linguistic-based rules
in combination with the reuse of existing
terminologies, % , then introducing
\textit{i.e.} exogenous disambiguation. 
% The tool we implemented exploits testified
% terms  at different steps of the process: chunking,
% parsing and extraction of term candidates.
Experiments reported here show that the combination of the two strategies allows the
extraction of a greater number of term candidates with a higher level of reliability. 
We further describe the extraction process involving both endogenous
and exogenous 
disambiguation  implemented in the term extractor \YaTeA. 
\end{abstract}

\section{Introduction}
Identifying and extracting terms from texts is now a well-known and widely
explored step in the terminology building process. Different
strategies can be applied: term extraction based on lexico-syntactic
markers \cite{Daille2003}, chunking based syntactic frontiers and
endogenous parsing % \cite{LEXTER}
\cite{Bourigault93}
, and distributional analysis
\cite{Bourigault&Fabre2000}. %  Later approaches proposed term candidate
% from parsed chunk.
% Different techniques for extracting terms from corpora have been
% studied and implemented in term extractors. They
%  use either morpho-syntactic patterns like ACABIT, % AJOUTER CITATION  
% syntactic frontiers like Lexter % AJOUTER CITATION
% or distributional analysis like Syntex. % AJOUTER CITATION
% In the latter, the list of the term candidates can be derived from the
% syntactic analysis of maximal noun phrases found in the corpus. 
% les techniques d'extraction sont maintenant assez développées et
% connues (utilisation de patrons morpho-syntaxique type ACABIT,
% utilisation de frontières syntaxiques type Lexter, analyse
% distribution type Syntex). Dans ce dernier cas, la liste de termes
% candidats peut-être calculée à partir de l'analyse syntaxique des
% groupes nominaux maximaux identifiés.
Those different techniques show satisfying extraction results
regarding the recall \cite{Cabre&al2001}. However, studying the
outputs of three term extractors applied to an English biomedical
corpus, we found that they are not adequate for highly technical texts
\cite{ExtraPloDocs3.1b}. The results of the extraction are generally
noisy for different reasons. First, some errors result from the
tagging of the corpus. The second limitation of such tools is their
difficulty to distinguish terms or variants from nominal phrases that
are not terms. Finally, they lack portability to new domains as it is
difficult to define parsing patterns large enough with a good
precision.

Extracting terms consists not only in identifying specific nominal
phrases but also in providing a reliable syntactic analysis. The latter
is commonly used to organise terminologies through a syntactic
network and to compute hierarchical relationships using lexical inclusion.  
%Those difficulties related to parsing add to those for chunking.
% Extraire des termes, c'est non seulement identifié des groupes
% nominaux précis mais aussi produire une analyse syntaxique correcte
% (permettant d'aller vers la structuration, par le réseau syntaxique
% et le calcul de relations hiérarchiques à l'aide de l'inclusion
% lexicale).
% S'ajoute donc aux problèmes précédents les problèmes de
% l'analyse syntaxique, les termes n'ayant pas forcément une structure
% classique.
% The main goal is thus to separate terms and variants from nominal
% phrases that are not valid terms.
Manually written rules based on linguistic clues are insufficient for
this task and must be combined with statistical methods.

Several strategies have been used and sometimes associated to finally
extract the term candidates: statistical filtering \cite{Daille2003}, manual filtering through the tool
interface  \cite{Bourigault93} or the exploitation of external
resources. %  We
% choose the latter solution by providing the term
% extractor with lists of testified terms.   
We propose a combination of the three methods.

% L'enjeu principal est maintenant de pouvoir distinguer parmi les
% termes candidats, les termes ou leurs variantes, des non termes.

% On peut exploiter plusieurs approches (pas forcément exclusives) :
% filtrage statistique (cf Acabit), filtrage manuel (lexter + interface)
% ou exploiter des connaissances extérieures (ce qu'on propose de faire).

The terminology extractor we implemented uses techniques comparable to
state-of-the-art tools, among which chunking based on morpho-syntactic
frontiers and production of the syntactic analysis of the terms extracted. 
% On a développé un outil d'extraction de termes qui se situe au niveau
% de l'état de l'art actuel dans ce domaine: utilisation de frontières
% morpho-syntaxiques, analyse syntaxique des termes.
We further propose new solutions for chunking and parsing %% . We
%% investigated the possibility
by using external resources%%  to augment
%% the quality of each of them
. In addition, we chose to perform positive
filtering in the  parsing step through the mechanism of islands of
reliability (see Section \ref{sec:endogenous}). In comparison, other tools
produce all parsing solutions and filter out non valid ones \textit{a posteriori}.

We first discuss the limitations of matching existing
terminologies on corpora and of automatic extraction tools. %  that suggest a need to tune lists of terms
% according to the working corpora.
As an answer to this, we propose a combination
of terminology extraction with the exploitation of testified
resources. 
We describe the extraction process of \YaTeA that implements the
method we propose. We finally present the results of experiments run on a
biomedical corpus to characterise the effects of recycling existing
terminologies in a term extractor.

\section{Which approach to identify terms?}
\label{sec:background}

%
% Plusieurs types de travaux : (1) Evaluation de l'utilisabilité et la
% couverture de corpus par des  ressources (McCray&al2002, Chute&al96)
% et indirectement (Bodenreider&al2002b)
%                              (2) Projection de une ou plusieurs
% ressources sur UMLS (Bodenreider&al2002 a et b)
%                              (3) Utilisation de la variation
% (Jacquemin96 a et b, Demetriou and Gaizauskas 2002),  Utilisation de
% Specialist (McCray&al94, Rindflesch&al1999)
%

%
% XXX distinction terms and named entities

Terms can be identified in corpora regarding two approaches~: matching
terms issued from terminological resources, or designing automatically
term extraction methods.

Using terminological resources to identify terms in texts addresses the
question of the usability of resources on working corpora, namely
their coverage and their adequacy. This leads to evaluate how terms
issued from resources, i.e. testified terms, match in the working
corpus.
As terminological resources are widely available in the biomedical
field, many experiments have been done on recycling terminologies to
identify terms in medical and biological corpora.
Coverage is generally mitigated.
The coverage of well-known classifications as \textsf{ICD-9},
\textsf{ICD-10} or \textsf{SNOMED III} have been observed on a 14,247
word corpus of clinical texts \cite{Chute&al96}. The evaluation leads
to conclude that no classification covers sufficiently the corpus,
although \textsf{SNOMED} has the better content coverage.
Similar observations have been noted regarding the evaluation of the
usability of \textsf{Gene Ontology} for NLP \cite{McCray&al2002}.
37\% of the \textsf{GO} terms are found in a 400,000 Medline citation
corpus. Results vary depending on the \textsf{GO} categories from 28\%
to 53 \% in the Medline corpus.  \cite{McCray&al2002} consider that
this low content coverage could be due to the size of the working
corpus or its narrow scope.
Still, content coverage is even worse on a set of 3 million
randomly selected noun phrases among 14 million terms extracted from
the Medline corpus \cite{Bodenreider&al2002b}: most of them are not
present in UMLS.
In \cite{Hamon2005}, we showed that, in the context of the indexation
of specialized texts, even if the combination of resources is useful
to identify numerous testified terms or variants, the indexation varies
greatly according to the documents.

% Terminological resources are widely used in specialized
% NLP tasks as a bootstrap and a help for several tasks or further
% linguistic analysis (parsing, ontology population, etc.). 
% %
% However, their relevance for the working corpus, namely their coverage
% and their adequacy, still remains a problem. It is needed to know how
% to select the most adequate terminology(ies) or how to tune them to
% the working corpora\footnote{See for instance the SNOMED tuning in
%   \cite{Nazarenko&al2000}.}.
%

Alternatives, based on the automatic extraction of terms, have been
widely proposed since the 90's. \cite{Cabre&al2001} give an overview of
the proposed term extractors. These term identification methods
generally exploit linguistic information like boundaries or, more often,
patterns. Such approaches are difficult to evaluate without a golden
standard and evaluations vary according to the methods. However, the
recall is generally good (\cite{Bourigault93} estimates the silence to
5\%), while the precision is rather low (\cite{Bourigault93} rejects
50\% of the extracted term candidates, the system discussed in \cite{Enguegard&al92b} 
has an error rate of 20\%).

Pure term extraction methods rarely use terminological
resources. Such domain information is rather exploited at the
filtering step \cite{Enguegard&al92b}. 
However, the usefulness of terminological resources in a term
extraction process is demonstrated in FASTR \cite{Jacquemin&al97}.
Results of this term variant extraction system are rather good as term
variation acquisition increases the terminological resource coverage. The limitation
of this approach is the acquisition of terms unrelated to testified
ones.

Regarding the works discussed above, it seems obvious that terminological resources
provide precious information that must be used in a term
identification task. However, exploiting terminological resources
requires their availability and adequacy on the targeted corpus. On the opposite, automatic term
extraction approaches suffer from a necessary human validation step. 
In that respect, we aim at combining both approaches by developing a
term extraction method that exploits terminological resources when
available.

% . %% This information can be used
%% at the chunking and the term parsing steps to take into specificity of
%% domain language and lakes of our approach.

% resource + corpus

\section{Strategy of term extraction}
\label{sec:endog-term-extr}

%%   The software is designed to identify noun phrases that look like
%%   terms in a corpus of sentences. In addition, it provides their
%%   syntactic analysis in a head-modifier format. This syntactic
%%   information will be used in the syntactic analysis of the sentences
%%   by simplifying terms by their head.

%%%%%%%%%%%%%%%%%%%%%%%%%%%%%%%%%%%%%%%%%%%%%%%%%%%%%%%%%%%%%%%%%%%%%%%%

The software \YaTeA, developed in the context of the ALVIS\footnote{European Project STREP
  IST-1-002068-STP,http://www.alvis.info/alvis/} project, aims at extracting noun phrases that look like
terms from a corpus. It provides their syntactic analysis
in a head-modifier format. 
As an input, the term extractor requires a corpus which has been
segmented into 
words and sentences, lemmatized and tagged with part-of-speech (POS)
information. The implementation of this term extractor allows to process large corpora.
%% The input data is a corpus annotated with
%% part-of-speech (POS) tags and lemmas. 
%% It takes a morpho-syntactically
%% tagged corpus as an input and exploits configuration files defined by
%% the user.
It is  not dependent on a specific language
in the sense that all linguistic features can be modified or created
for a new language, sub-language or tagset. % This allows to use any
% part-of-speech tagset.
In the experiments described here, we used the
\textsc{genia} 
tagger\footnote{http://www-tsujii.is.s.u-tokyo.ac.jp/GENIA/tagger/} \cite{Tsuruoka&al2005}
which is specifically designed for
biomedical corpora and uses the Penn TreeBank tagset.

%The tool is implemented in Perl. 

The main strategy of analysis of the term candidates is
based on the exploitation of simple parsing patterns and endogenous disambiguation. Exogenous disambiguation is also made
possible for the identification and the analysis of term candidates by the use of external resources, \textit{i.e.} lists of
testified terms. 

This section includes the presentation of  both endogenous and
exogenous disambiguation strategies. We also describe the whole
extraction process implemented in \YaTeA.

%%%%%%%%%%%%%%%%%%%%%%%%%%%%%%%%%%%%%%%%%%%%%%%%%%%%%%%%%%%%%%%%%%%%%%%%

%% The term extractor aims at identifying and extracting noun phrases
%% which are potential terms ({\em i.e. } term candidates). Moreover each
%% term candidate is syntactically analysed in order to identify head and
%% modifier components. In that respect, it provides the syntactic head
%% of the term which is used in the syntactic parse step of the whole NLP
%% chain. Before parsing, terms identified in the corpus are simplified by their
%% syntactic head in order to decrease the complexity of the sentences.

%% We plan to set a confidence degree to each step of the term
%% analysis. 

\subsection{Endogenous and exogenous disambiguation}
\label{sec:endogenous}

Endogenous disambiguation consists in the exploitation of intermediate
%% chunking and parsing
extraction 
results for the parsing of a given Maximal Noun Phrase
(MNP). 

All the MNPs
corresponding to parsing patterns %  or testified terms
are
parsed first. In a second step, remaining unparsed MNPs are processed
using the MNPs parsed during the first step as \emph{islands of reliability}. 
An \emph{island of reliability} is a subsequence
(contiguous or not) of a MNP that corresponds %% This subsequence
to % a testified term or
a shorter term candidate in either its inflected or lemmatized form. %%  that was
%% parsed during the first step of the parsing process
 It is used as an anchor as follows: %% This subsequence along with its internal analysis is used as
%% an anchor in the parsing of the MNP.
%% Islands are used to simplify the
%% POS sequence of the MNP for which no parsing pattern was found.
the subsequence covered by the island is 
reduced to the word found to be the syntactic head of the island. Parsing patterns
are then applied to the simplified MNP. 

\begin{figure}[!htbp]
\centering
  %% \textbf{differential pattern of cot gene expression}\\
%%   \textbf{JJ NN of NN NN NN}\\
%%   ISLANDS: \\
%%   - cot gene : head = gene\\
%%   - gene expression : head = expression\\
%%   REDUCED POS SEQUENCE:   JJ NN of NN\\
%%   PARSE using islands:  ( ( differential  pattern ) of  ( ( cot gene )
%%   expression ) ) \\
%%    PARSE without using islands: *( differential ( pattern of ( cot
  %%    ( gene  expression ) )\\
  %\includegraphics[scale=0.6]{endogenous.eps}
   \includegraphics[scale=0.6]{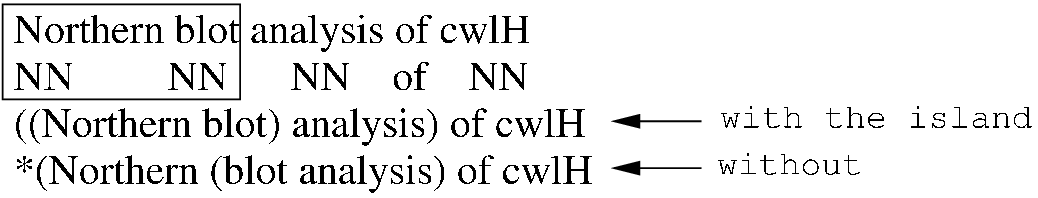}
  \caption{Effect of an island on parsing}
  \label{fig:endogenous}
\end{figure}

This feature
allows the parse of complex noun phrases using a limited number of
simple parsing patterns (80 patterns containing a maximum of 3
content words were defined for the experiments described below). 
In addition,   
% As a consequence,
islands increase the degree of reliability of the
parse as shown in Figure \ref{fig:endogenous}. 

%% The identification and the syntactic analysis of the terms is similar
%% to those used in the French term extractor Lexter \cite{lexter}:
%% identification of maximal noun phrases, analysis in head and modifer
%% of the noun phrase components. [J'AIME PAS TROP CETTE PHRASE]
%% The implementation of this term extractor allows to process large corpora.
%% The tool can be adapted to the corpus and a language thanks to the
%% configurable files defining the chunking and analysis of the input
%% data. In that respect, any part-of-speech tagset can be used.
%% We implement an endogenous  syntactic disambiguation of the terms
%% which uses  clues in the corpora to produce the best parse(s) for a term.
%% Exogeneous disambiguation is also made possible by using existing
%% terminologies (\textit{i.e.} testified terms) as an input of the extractor.
%% An index of reliability is assigned to each parse.
%%%%%%%%%%%%%%%%%%%%%%%%%%%%%%%%%%%%%%%%%%%%%%%%%%%%%%%%%%%%%%%%%%%%%%%%

%\subsection{Exogenous disambiguation}
%\label{sec:exogenous}
\YaTeA allows exogenous disambiguation,  \textit{i.e.} the exploitation of existing (testified) terminologies to
assist the chunking, parsing and extraction steps. 

During chunking, sequences of words corresponding to testified terms
are identified. They cannot be further split or deleted. Their POS tags and
lemmas can be corrected according to those associated to the testified term. 
If an MNP corresponds to a testified term for which a parse exists
(provided by the user or computed using parsing patterns), it is
recorded as a term candidate with the highest score of reliability.
Similarly to endogenous disambiguation, subsequences of MNPs
corresponding to testified terms are % used to simplify the POS sequence
% of the MNP
used as islands of reliability in order to augment the number and
quality of parsed MNPs%  (Figure \ref{fig:exogenous})
.

% \begin{figure}[!htbp]
% \centering 
%   \includegraphics[scale=0.6]{exogenous.eps}
% %% \begin{small}
% %%   \textbf{Northern blot analysis of cwlH}\\
% %%   \textbf{NN NN NN of NN}\\
% %%   ISLANDS from testified: \\
% %%   - northern blot : head = blot\\
% %%   REDUCED POS SEQUENCE: NN NN of NN  \\
% %%   PARSE using islands: ( ( ( Northern blot ) analysis ) of cwlH )\\
% %%    PARSE without using islands: *( (  Northern ( blot  analysis ) ) of cwlH )\\
%  \caption{Exogenous disambiguation for parsing}
%  \label{fig:exogenous}
% %% \end{small}
% \end{figure}

\subsection{Term candidate extraction process}\label{sec:term-extr-proc}
A noun phrase is extracted from the corpus and considered a term candidate if at least one
parse is found for it. This is performed in three main steps,
(1) \emph{chunking}, \textit{i.e.} construction of a list of Maximal Noun
Phrases from the corpus, (2) \emph{parsing}, \textit{i.e.} attempts
to find at least one syntactic parse for each MNP and, 
(3) \emph{extraction} of term candidates.
The result of the term extraction process is two lists of noun phrases:  one
contains parsed MNPs, called \emph{term candidates}, the other
contains MNPs for which no parse was found. Both lists are proposed to
the user through a validation interface (ongoing development).

%  We assume that the terms are
% part of the parsed maximal noun phrases and thus consider each
% internal node of the syntactic tree of a MNP as a
% potential term, i.e. a term candidate.

% Linguistic data defined by the user are used in the chunking and parsing steps as well as
% existing terminologies when provided.

% The term extractor has three steps: (1) the chuncking (identification of
% the maximal noun phrases), (2) the analysis of the term candidates fully
% covered by at least one parsing pattern (defined maximal noun phrases), and (3) the
% analysis of the remaining terms (under-defined  maximal noun phrases)
% using the results of step 2.

% It requires several types of information:
% \begin{itemize}
% \item chunking frontiers, used to identify the maximal noun phrases;

% \item cleaning frontiers, used to remove words (\emph{e.g.} determiners) that
%   cannot be in a start or final position of a term;

% \item chunking exceptions, used to prevent the corpus
%   from chunking in some specific cases;

% \item forbidden structures, used to identifiy the word or 
%   tag sequences that cannot appear in a term;

% \item parsing patterns.

% \end{itemize}

%\paragraph{Description of the extraction steps}~\\

\begin{enumerate}

\item \textbf{Chunking}: the corpus is chunked into Maximal
  Noun Phrases. %% as shown in bold in the following examples. 
%%   \begin{figure}
%%   To understand the \textbf{role of the pro sequence} in controlling
%%   \textbf{sigmaK activity}, we have constructed \textbf{NH2-terminal
%%     truncations of pro-sigmaK} and characterized their behavior in
%%   vitro at the \textbf{gerE promoter}.\\
%%   \textbf{Bacillus subtilis Spo0A} activates \textbf{transcription} from both \textbf{sigmaA-} and \textbf{sigmaH-dependent promoters}.
%% \end{figure}

  The POS tags associated to the words of the corpus are used to delimit the MNPs
according to the resources provided by the user: chunking
frontiers and exceptions, forbidden structures and potentially,
testified terms. 

\emph{Chunking frontiers} are tags or words that are not allowed to
  appear in MNPs, e.g. verbs (VBG) or prepositions (IN). \emph{Chunking exceptions} are
used to refine frontiers. For instance, \textit{"of"} is a
  frontier exception to prepositions,  \textit{"many"} and
  \textit{"several"} being exceptions to adjectives. \emph{Forbidden
    structures} %  stand as 
  are 
  exceptions for more complex structures and are used to prevent from
  extracting sequences that look like terms (syntactically valid) but
  are known not to be terms or parts of terms like \textit{"of course"}.
  MNPs and subparts of MNPs corresponding to testified terms (when available) are
  protected and cannot be modified using the chunking data. For
  instance, the tag FW is \textit{a priori} not allowed in MNPs. However, if an
  MNP is equal to or contains the testified term \textit{"in/IN vitro/FW"}, it will be kept as such.
\item \textbf{Parsing}: for each identified MNP type, except
  monolexical MNPs,
  different parsing methods are applied in decreasing order of
  reliability. Once a method succeeds in parsing the MNP, the parsing
  process comes to an end. Still, one method can compute several
  parses for the same MNP, making the parsing non-deterministic if desired. 
  We consider 3 different parsing methods:
  \begin{itemize}
  \item \textsc{tt-covered}: the MNP inflected or lemmatized form
    corresponds to one or several combined  testified terms (TT);
  %% \item \textsc{tt-combination}: the MNP inflected or lemmatized form
%%     corresponds to the combination of several testified terms (TT);
  \item \textsc{pattern-covered}: the POS sequence of the (possibly simplified) MNP corresponds to a
    parsing pattern provided by user;
  \item \textsc{progressive}: the MNP is progressively reduced at its left and
    right ends by the application of parsing patterns. Islands of
    reliability from term candidates or testified terms are also used to reduce the MNP sequence of the MNP to allow the application of parsing patterns.
  \end{itemize}
\item \textbf{Extraction} of term candidates: MNPs that received a
  parse in the previous processing step are considered term
  candidates. Statistical measures will further be implemented to
  order MNPs according to their likelihood to be a term in order to
  facilitate their validation by the user. 
\end{enumerate}
%% The maximal noun phrases that remain unparsed are moved
%% appart. We consider that those are illformed chunks that cannot
%% contain portential terms.
% \section{Input format}
% \label{sec:input-format}
% \subsection{Corpus}
% As an input, the term extractor requires a corpus which has been
% segmented in words and sentences, lemmatized and tagged with
% part-of-speech (POS)
% The tool is implemented in Perl. It takes a morpho-syntactically
% tagged corpus as an input and exploits configuration files defined by
% the user.
\section{Experiments}
\label{sec:experiments}
To characterise the effects of resources on term extraction,
we compare %% two versions of \YaTeA on biomedical data. \texttt{no resource}
%% is the result for \YaTeA alone while \texttt{GO+MeSH} is the result for \YaTeA exploiting lists
%% of terms from two biomedical resources. We present the effects of
%% those resources on
%% chunking, parsing and extraction of term candidates.  
the results provided by \YaTeA using or not existing
terminologies on a biomedical corpus. We present and comment the effects on
 chunking, parsing and extraction of the term candidates.

\subsection{Materials}
\label{sec:materials}

\subsubsection{Working corpus}
\label{sec:working-corpus}

%% We carry out a preliminary evaluation of the term extraction
%% performance on a corpus of 932 sentences\footnote{Experiments on a
%%   larger corpus will be done for the final version of this paper}. Those sentences have been
%% selected from 2~209 Medline abstract describing genomic interaction of
%% the model organism \textit{``Bacillus subtilis''}. This corpus contains 33~429
%% words.
We carry out an experiment on a corpus of 16,600 sentences (438,513 words)
describing genomic interaction of the model organism
\textit{``Bacillus subtilis''}. The corpus was tagged and lemmatized
using the \textsc{genia} tagger \cite{Tsuruoka&al2005}. %% This corpus contains 438~513
%% words.

\subsubsection{Terminological resources} 
\label{sec:term-reso}

To study the reuse of terminologies in the term extractor,
we tested two types of resources: terms from two public databases and a
list of terms extracted from the working corpus.
We first selected and merged two specialized resources covering genomic vocabulary: Gene
Ontology \cite{GOConsortium2001b} and MeSH \cite{MeSH}, both 
issued from the december 2005 release of UMLS \cite{UMLS}.
The \textsf{Gene Ontology
  resource}\footnote{\url{http://www.geneontology.org/}} (henceforth
  GO) aims at proposing a controlled vocabulary related to the genomic
  description of any organism, prokaryotes as well as eukaryotes
  \cite{GOConsortium2001}. %% GO proposes a list of 24,803 terms......
%%  The \textsf{Gene Ontology
%%   resource}\footnote{\url{http://www.geneontology.org/}} (henceforth
%% GO) gathers the vocabulary that can be applied to the eukaryotes
%% \cite{GOConsortium2001}. It has been built from three model organism
%% databases~: Saccharomyces Genome Database (SGD), Drosophila genome
%% database (Flybase), Mouse Genome Informatics databases (MGD/GXD). 
%% Although Bacillus subtilis is a prokaryote, GO aims at unifying the
%% biological information. In that respect, we consider how GO terms
%% match our working corpora.
GO proposes a list of 24,803 terms. The \textsf{Medical Subject Headings}
thesaurus\footnote{\url{http://www.nlm.nih.gov/mesh}} (henceforth
\textsf{MeSH}) is dedicated to the indexation of the Medline database.
% It gathers terms used in the medical domain \cite{Cote96}.
The UMLS version
of the MeSH offers 390,489 terms used in the medical domain \cite{Cote96}.

The TAC (Terms Acquired in Corpus) resource is a list of 515 terms
extracted from our working corpus using three term extractors
\cite{ExtraPloDocs3.1b}. The 515 terms occur at least 20 times in the
corpus and were validated by a biologist.

\subsection{Results}\label{sec:experiments-results}
We present and comment the results of \YaTeA using no resource,
the combination of GO and MeSH (GO+MeSH) and finally the TAC resource.

\subsubsection{Chunking} is affected by resources in several ways. 
%The effects of the resources on chunking are characterized by the
%results presented
As shown in Table \ref{tab:chunking}, they allow the identification of
new MNPs that were originally rejected due to their POS tag(s). 
%Results presented in Table \ref{tab:chunking} first show that using existing terminologies leads
%to the identification of a greater number of MNPs. 
In addition, the MNPs tend to be longer and monolexical terms less numerous. 
 As MNPs are more complex, the number of types of POS sequences to be
 parsed is augmented. However, this increase in diversity is
expected to be compensated by the parsing mechanism related to islands of reliability.  
\begin{table}[!htbp]
  \centering
  \caption{Effects of resources on chunking}\label{tab:chunking}
  \begin{tabular}{|c||c|c||c|c||c||c|}
    \hline
    Version & \multicolumn{2}{|c||}{MNPs} &
    \multicolumn{2}{|c||}{Monolexical} & Words/ &
    POS sequences \\
    \cline{2-5}
            & types & occ & types & occ & complex MNP & types \\
\hline
% \texttt{no resource} &4~065  &6~691  &938  &2~637  & 3.51 & 510 & 0.12 \\
% \texttt{GO+MeSH} &  4~088  &6~691  & 932 & 2~590 & 3.53  & 565 &  0.14\\
\texttt{no resource} & 45,716 &84,810  & 6,989 & 30,815 &3.61  &2,965\\
\texttt{GO+MeSH} &  46,079  & 85,004 & 6,949 & 30,272  & 3.63 &3,256 \\
\texttt{TAC} & 46,315 & 84,918 &6,934 &29,695 &3.65 & 3,500\\
    \hline
  \end{tabular}
  
\end{table}
%  \begin{tabular}{|c||c|c||c|c||c|c|}
%%  \begin{tabular}{|c|c|c|}
%% \hline
%%  & &
%% Go+MeSH\\% &
%% Method & no resources &correction \\
%% %\multicolumn{2}{|c|}{Go+MeSH no correction}\\
%% \hline
%% MNPs                &     &    \\%   &       &      \\
%% Monolexical         &        &   \\%  &       &      \\
%% Words/complex MNP   &     &  \\%     &      \\
%% POS types           &     & \\%    &       &  %    \\
%% POS diversity       &     &  \\%    &       &  %    \\
%% \hline
%%   \end{tabular}
%%   \caption{Chunking}
%% \end{table}
%
%
%
% \begin{table}[!htbp]
%   \centering
%   \begin{tabular}{|c||c|c|c||c|c|}
%     \hline 
%      & \multicolumn{3}{|c||}{MNPs} & \multicolumn{2}{|c|}{MNP length}\\
%     \cline{2-6} 
%     Version & types & occ & diversity &  types & occ \\
%     \hline 
%     no resources  &  &  &  & & \\
%     Go+MeSH  &  &  &  & & \\
%     Go+MeSH no correction &  &  &  & & \\
%     \hline 
%   \end{tabular}
%   \caption{Chunking}
% \end{table}
%
%
%\paragraph{Without resource}
%
%même chunking pour les deux étiqueteurs (curieux)
\subsubsection{Parsing MNPs} 
\begin{figure}[!htbp]
      \centering
      \includegraphics[scale=0.5]{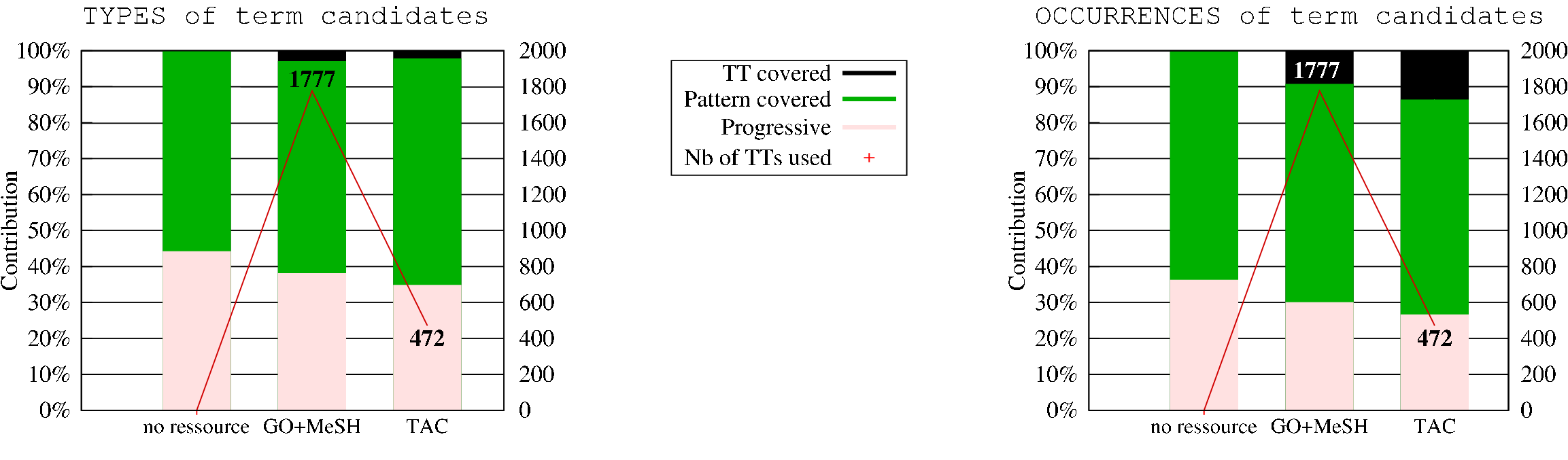}
      \caption{Contribution of parsing methods}
      \label{fig:parsing}
    \end{figure}

is %% the process step expected to be the
%% most 
also affected by the use of resources that increase the 
    reliability of parses since testified terms are used as islands of
    reliability. The contribution of each parsing method is presented in Figure
    \ref{fig:parsing} regarding the total types and occurrences of MNPs. %%  in decreasing order of reliability and their contribution in
%%     percentages is expressed in terms of types and occurences. 
    Interestingly, the  \textsc{tt-covered} method discharges the \textsc{progressive} method which is the least
    reliable. The increase in the contribution of the \textsc{pattern-covered}
    method is explained by the extraction of new short terms like
    species names, e.g. \textit{``Escherichia/FW coli/FW''}, the
    expansion of monolexical terms like \textit{"DNA/NNP"}
    to \textit{"DNA/NNP binding/VBG"} that results from tag correction
    (VBG replaced by NN) and the simplification of MNPs using islands
    before the application of the parsing patterns. % Such terms are
%     very frequent in our corpus but are ignored by the \texttt{no resource}
%     version for not containing at least a noun tag.

The comparison of the diagrams on types and occurrences
shows that both
resources cover frequent terms. Still, GO+MeSH unsurprisingly contributes little
(1777 terms out of 415,292 are used) compared to the corpus-tuned 
resource (TAC). % This shows the interest of using terminologies extracted
% from corpora as a bootstrap for the extraction of new terms.

\subsubsection{Extraction of term candidates} is dependent on both
preceding steps as an MNP found during chunking is considered a term
candidate if at least one parse is found for it. Statistical filtering methods,
that will be further implemented, are expected to provide
qualitative information on term candidates and to allow the
extraction of monolexical terms. 
On a quantitative point of view, using existing terminologies
results in the extraction of a greater number of term candidates.

%% We found that the recycling of existing terminologies allows the
%% acquisition of new term candidates (29 types).
%% As \YaTeA involves
%% endogenous disambiguation, the proportion of new terms is expected to augment on larger corpora. 
%% A slight decrease ($<$1\%) in the part of 
%% parsed MNPs indicates that new terms are complex or show specific
%% patterns. This can be countered by adapting parsing patterns or
%% providing the syntactic analysis for testified terms, which is allowed
%% by \YaTeA but was not done in the context of this first experiment. 

%% \begin{table}[!htbp]
%%   \caption{Extraction of complex term candidates}\label{tab:extraction}
%%   %\begin{tabular}{|c||c|c||c|c||c|c|}
%%   \centering
%% \begin{tabular}{|l||c|c||c|c|}
%% \hline
%% Method &\multicolumn{2}{|c||}{no resources} &
%% \multicolumn{2}{|c|}{Go+MeSH}\\% &
%% %\multicolumn{2}{|c|}{Go+MeSH no correction}\\
%% %\cline{2-7}
%% \cline{2-5}
%%                 & types & occ & types & occ \\%& types & occ \\
%% \hline
%% Complex MNPs    &  3127   & 4054 & 3156 & 4101 \\%  &       &
%% Term Candidates  & 97.03\%   & 97.66\%& 95.98\% & 96.63\% \\%  &       &      \\
%% Rejected         & 2.97\%   & 2.34\% & 4.02\%  & 3.36\%  \\%  &      \\
%% \hline
%%   \end{tabular}\\
%% \end{table}

%% \subsection{Discussion}
%% \label{sec:discussion}

\section{Conclusion and future works}
Term extractors on the one hand and terminology matching techniques on
the other hand show limitations in term acquisition and term exploitation
respectively. To both reduce noisy results of the extraction and augment the
coverage of existing terminologies, we proposed to combine both
techniques in a term extractor. 
With a first experiment on a biomedical corpus, we showed that the
exploitation of existing terminologies in a term extractor positively influences
 the identification of maximal noun phrases, their parsing and finally the
 extraction of lists of term candidates. 
The result of the extraction is a corpus-tuned list of term
candidates. 
It is composed of a subset of the external resource(s) augmented with term
candidates acquired in the corpus in conformity with the former. 
As future works, we intend to add statistical features to assist the endogenous and
exogenous disambiguation. The handling of coordinations is also about
to be integrated. Finally, a precise evaluation of the outputs of
\YaTeA through a validation interface % will be held in the coming          % months
is planed.
\label{sec:conclusion}


\bibliography{th-biblio}
\bibliographystyle{splncs}

\end{document}